\documentclass[11pt,a4paper]{article}
\usepackage[hyperref]{acl2021}
\usepackage{times}
\usepackage{latexsym}

\usepackage{booktabs}
\usepackage{graphicx}
\usepackage{subfig}
\usepackage{textcomp}
\usepackage{multirow}
\usepackage{bm}

\usepackage{microtype}

\aclfinalcopy 

\newcommand{\approximately}{{\raise.17ex\hbox{$\scriptstyle\sim$}}}

\title{ViTA: Visual-Linguistic Translation by Aligning Object Tags}

\author{Kshitij Gupta \qquad Devansh Gautam \qquad Radhika Mamidi \\
  International Institute of Information Technology Hyderabad \\
  \texttt{\{kshitij.gupta,devansh.gautam\}@research.iiit.ac.in},\\ \texttt{radhika.mamidi@iiit.ac.in}
}

\date{}

\begin{document}
\maketitle
\begin{abstract}

Multimodal Machine Translation (MMT) enriches the source text with visual information for translation. It has gained popularity in recent years, and several pipelines have been proposed in the same direction.
Yet, the task lacks quality datasets to illustrate the contribution of visual modality in the translation systems.
In this paper, we propose our system under the team name \textit{Volta} for the Multimodal Translation Task of WAT 2021\footnote{\url{http://lotus.kuee.kyoto-u.ac.jp/WAT/evaluation/index.html}}~\citep{nakazawa-etal-2021-overview} from English to Hindi. We also participate in the textual-only subtask of the same language pair for which we use mBART, a pretrained multilingual sequence-to-sequence model.
For multimodal translation, we propose to enhance the textual input by bringing the visual information to a textual domain by extracting object tags from the image.
We also explore the robustness of our system by systematically degrading the source text. Finally, we achieve a BLEU score of 44.6 and 51.6 on the test set and challenge set of the multimodal task.
\end{abstract}

\section{Introduction}

Machine Translation deals with the task of translation between language pairs and has been an active area of research in the current stage of globalization.
In the task of multimodal machine translation, the problem is further extended to incorporate visual modality in the translations. The visual cues help build a better context for the source text and are expected to help in cases of ambiguity.

With the help of visual grounding, the machine translation system has scope for becoming more robust by mitigating noise from the source text and relying on the visual modality as well.

In the current landscape of multimodal translation, one of the issues is the limited datasets available for the task. Another contributing factor is that often the images add irrelevant information to the sentences, which may act as noise instead of an added feature. The available datasets, like Multi30K~\citep{W16-3210}, are relatively smaller when compared to large-scale text-only datasets~\citep{DBLP:journals/corr/BahdanauCB14}. The scarcity of such datasets hinders building robust systems for multimodal translation.

To address these issues, we propose to bring the visual information to a textual domain and fine-tune a high resource unimodal translation system to incorporate the added information in the input. We add the visual information by extracting the object classes by using an object detector and add them as tags to the source text. Further, we use mBART, a pretrained multilingual sequence-to-sequence model, as the base architecture for our translation system. We fine-tune the model on a textual-only dataset released by \citet{kunchukuttan-etal-2018-iit} consisting of 1,609,682 parallel sentences in English and Hindi. Further, we fine-tune it on the training set enriched with the object tags extracted from the images. We achieve state-of-the-art performance on the given dataset. The code for our proposed system is available at \url{https://github.com/kshitij98/vita}.

The main contributions of our work are as follows:

\begin{itemize}
    \item We explore the effectiveness of fine-tuning mBART to translate English sentences to Hindi in the text-only domain.
    \item We further propose a multimodal system for translation by enriching the input with the object tags extracted from the images using an object detector.
    \item We explore the robustness of our system by a thorough analysis of the proposed pipelines by systematically degrading the source text and finally give a direction for future work.
\end{itemize}

The rest of the paper is organized as follows. 
We discuss prior work related to multimodal translation.
We describe our systems for the textual-only and multimodal translation tasks.
Further, we report and compare the performance of our models with other systems from the leaderboard.
Lastly, we conduct a thorough error analysis of our systems and conclude with a direction for future work.

\section{Related Work}

Earlier works in the field of machine translation largely used statistical or rule-based approaches, while neural machine translation has gained popularity in the recent past. \citet{kalchbrenner-blunsom-2013-recurrent} released the first deep learning model in this direction, and later works utilize transformer-based approaches \citep{NIPS2017_3f5ee243, song2019mass, lample2019cross, edunov-etal-2019-pre,  liu2020multilingual} for the problem.

Multimodal translation aims to use the visual modality with the source text to help create a better context of the source text. \citet{specia-etal-2016-shared} first conducted a shared task on the problem and released the dataset, Multi30K~\citep{W16-3210}. It is an extended German version of Flickr30K~\citep{young-etal-2014-image}, which was further extended to French and Czech~\citep{elliott-etal-2017-findings, barrault-etal-2018-findings}. For multimodal translation between English and Hindi, \citet{hindi-visual-genome:2019} propose a subset of Visual Genome dataset~\citep{10.1007/s11263-016-0981-7} and provide parallel sentences for each of the captions. 

Although both English and Hindi are spoken by a large number of people around the world, there has been limited research in this direction. \citet{dutta-chowdhury-etal-2018-multimodal} created a synthetic dataset for multimodal translation of the language pair and further used the system proposed by \citet{calixto-liu-2017-incorporating}. Later, \citet{sanayai-meetei-etal-2019-wat2019} work with the same architecture on the multimodal translation task in WAT 2019. \citet{laskar-etal-2019-english} used a doubly attentive RNN-based encoder and decoder architecture~\citep{calixto-liu-2017-incorporating,calixto-etal-2017-doubly}. \citet{laskar-etal-2020-multimodal} also proposed a similar architecture and pretrained on a large textual parallel dataset \citep{kunchukuttan-etal-2018-iit} in their system.

\section{System Overview}

In this section, we describe the systems we use for the task.

\subsection{Dataset Description}

\begin{table}
\centering
\resizebox{\columnwidth}{!}{%
\begin{tabular}{lrrrr}
\toprule
                     & \textbf{Train} & \textbf{Valid} & \textbf{Test} & \textbf{Challenge} \\ \midrule
\#sentence pairs & 28,930 & 998 & 1,595 & 1,400 \\
Avg. \#tokens (source) & 4.95 & 4.93 & 4.92 & 5.85 \\
Avg. \#tokens (target) & 5.03 & 4.99 & 4.92 & 6.17 \\
\bottomrule
\end{tabular}%
}
\caption{The statistics of the provided dataset. The average number of tokens in the source and target language are reported for all the sentence pairs.}
\label{tab:data_stats}
\end{table}

We use the dataset provided by the shared task organizers~\citep{hindi-visual-genome:2019}, which consists of images and their associated English captions from Visual Genome~\citep{10.1007/s11263-016-0981-7} along with the Hindi translations of the captions. The dataset also provides a challenge test which consists of sentences where there are ambiguous English words, and the image can help in resolving the ambiguity. The statistics of the dataset are shown in Table~\ref{tab:data_stats}. We use the provided dataset splits for training our models.

We also use the dataset released by \citet{kunchukuttan-etal-2018-iit} which consists of parallel sentences in English and Hindi. We use the training set, which contains 1,609,682 sentences, for training our systems.

\subsection{Model}

We fine-tune mBART, which is a multilingual sequence-to-sequence denoising auto-encoder that has been pre-trained using the BART~\citep{lewis-etal-2020-bart} objective on large-scale monolingual corpora of 25 languages, including both English and Hindi. The pre-training corpus consists of 55,608 million English tokens~(300.8 GB) and 1,715 million Hindi tokens~(20.2 GB). Its architecture is a standard sequence-to-sequence Transformer~\citep{NIPS2017_3f5ee243}, with 12 encoder and decoder layers each and a model dimension of 1024 on 16 heads resulting in \approximately680 million parameters. To train our systems efficiently, we prune mBART's vocabulary by removing the tokens which are not present in the provided dataset or the dataset released by \citet{kunchukuttan-etal-2018-iit}.

\subsubsection{mBART}




\begin{table*}[!t]
\centering
\resizebox{0.7\textwidth}{!}{
\begin{tabular}{lcccccc}
\toprule
\multirow{2}{*}[-2.5pt]{\textbf{Model}} & \multicolumn{3}{c}{\textbf{Test Set}} & \multicolumn{3}{c}{\textbf{Challenge Set}}\\
\cmidrule(lr){2-4}
\cmidrule(lr){5-7}
  & \textbf{BLEU} & \textbf{RIBES} & \textbf{AMFM} & \textbf{BLEU} & \textbf{RIBES} & \textbf{AMFM}\\ \midrule
\multicolumn{7}{l}{\textit{Text-only Translation}} \\ \midrule
CNLP-NITS-PP & 37.01 & 0.80 & 0.81 & 37.16 & 0.77 & 0.80\\
ODIANLP & 40.85 & 0.79 & 0.81 & 38.50 & 0.78 & 0.80\\
NLPHut & 42.11 & 0.81 & 0.82 & 43.29 & 0.82 & 0.83\\
mBART (ours) & \textbf{44.12} & \textbf{0.82} & \textbf{0.84} & \textbf{51.66} & \textbf{0.86} & \textbf{0.88}\\ \midrule
\multicolumn{7}{l}{\textit{Multimodal Translation}} \\ \midrule
CNLP-NITS & 40.51 & 0.80 & 0.82 & 33.57 & 0.75 & 0.79\\
iitp & 42.47 & 0.81 & 0.82 & 37.50 & 0.79 & 0.81\\
CNLP-NITS-PP & 39.46 & 0.80 & 0.82 & 39.28 & 0.79 & 0.83\\
ViTA (ours) & \textbf{44.64} & \textbf{0.82} & \textbf{0.84} & \textbf{51.60} & \textbf{0.86} & \textbf{0.88}\\ \bottomrule
\end{tabular}%
}
\caption{Performance of our proposed systems on the test and challenge set.}
\label{tab:results}
\end{table*}

We fine-tune mBART for text-only translation from English to Hindi and feed the English sentences to the encoder and decode Hindi sentences. We first fine-tune the model on the dataset released by \citet{kunchukuttan-etal-2018-iit} for 30 epochs, and then fine-tune it on the Hindi Visual Genome dataset for 30 epochs.

\subsubsection{ViTA}

We again fine-tune mBART for multimodal translation from English to Hindi but add the visual information of the image to the text by adding the list of object tags detected from the image.
We feed the English sentences along with the list of object tags to the encoder and decode Hindi sentences. For feeding the data to the encoder, we concatenate the English sentence, followed by a separator token `\#\#', followed by the object tags which are separated by `,'.  We use Faster R-CNN with ResNet-101-C4 backbone\footnote{We use the implementation available in Detectron2 (\url{https://github.com/facebookresearch/detectron2}).}~\citep{DBLP:conf/nips/RenHGS15} to detect the list of objects present in the image. We sort the objects by their confidence scores and choose the top ten objects.

For training the model, we first fine-tune the model on the dataset released by ~\citet{kunchukuttan-etal-2018-iit}. Since this is a text-only dataset, we do not add any object tag information. Afterward, we fine-tune the model on Hindi Visual Genome dataset, where each sentence has been concatenated with object tags. Initially, we mask \approximately15\% of the tokens in each sentence to incentivize the model to use the object tags along with the text and fine-tune the model on masked sentences along with object tags for 30 epochs. Finally, we train the model for 30 more epochs on Hindi Visual Genome dataset with unmasked sentences and object tags.

\subsection{Experimental Setup}

We implement our systems using the implementation of mBART available in the fairseq library\footnote{\url{https://github.com/pytorch/fairseq}}~\citep{ott-etal-2019-fairseq}. We fine-tune on 4 Nvidia GeForce RTX 2080 Ti GPUs with an effective batch size of 1024 tokens per GPU. We use the Adam optimizer~($\epsilon = 10^{-6}, \beta_{1} = 0.9, \beta_{2} = 0.98$)~\citep{adamoptimizer} with 0.1 attention dropout, 0.3 dropout, 0.2 label smoothing and polynomial decay learning rate scheduling. We validate the models every epoch and select the best checkpoint after each training based on the best validation BLEU score. To train our systems efficiently, we prune the vocabulary of our model by removing the tokens which do not appear in any of the datasets mentioned in the previous section. While decoding, we use beam search with a beam size of 5.

\section{Results and Discussion}

The BLEU score~\citep{papineni-etal-2002-bleu} is the official metric for evaluating the performance of the models in the leaderboard. The leaderboard further uses RIBES~\citep{isozaki-etal-2010-automatic} and AMFM~\citep{banchs-li-2011-fm} metrics for the evaluations.
We report the performance of our models after tokenizing the Hindi outputs using \texttt{indic-tokenizer}\footnote{\url{https://github.com/ltrc/indic-tokenizer}} in Table \ref{tab:results}.

\begin{figure}
    \centering
    \includegraphics[width=\columnwidth,keepaspectratio]{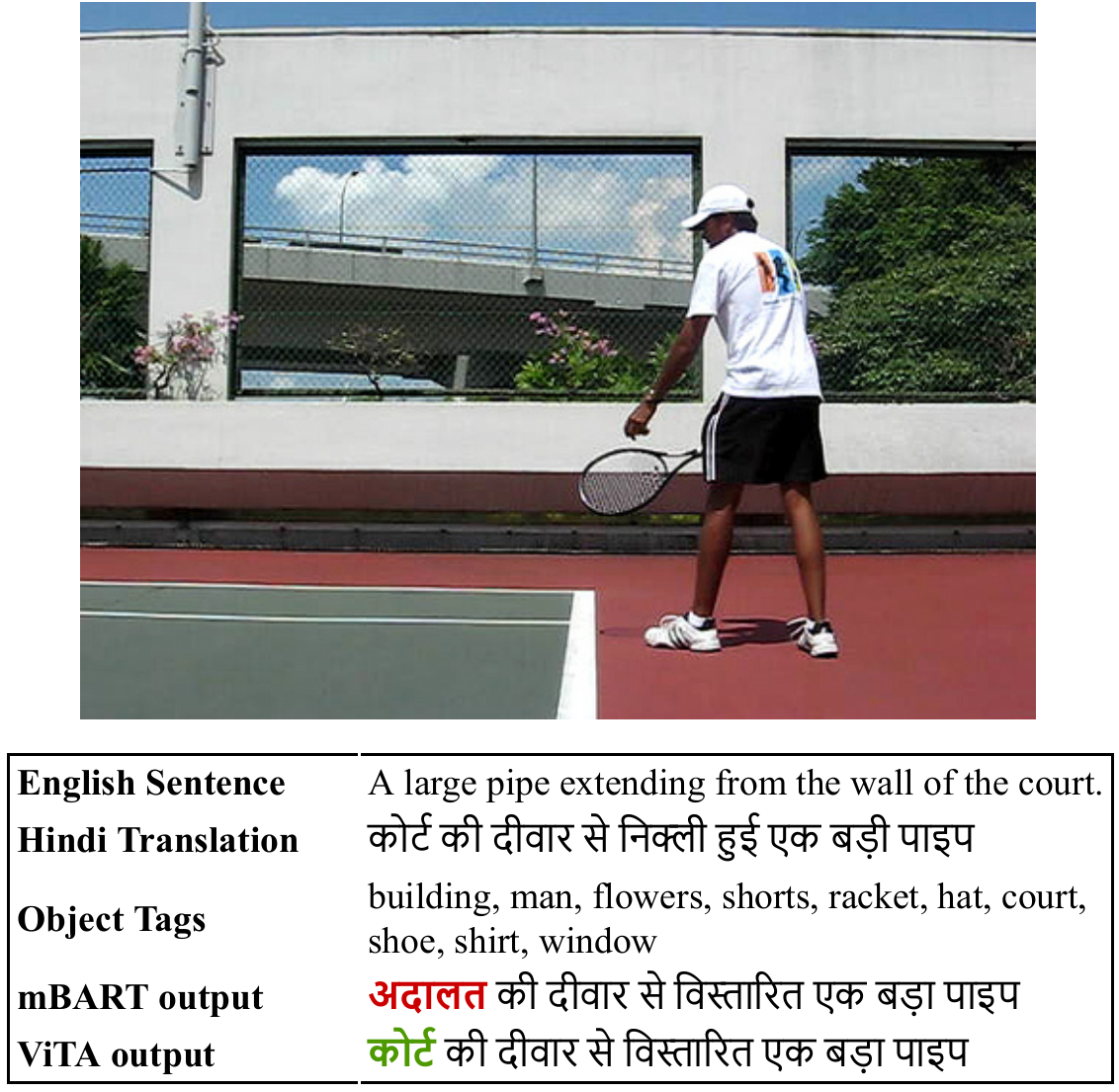}
    \caption{A translation example from the challenge set which illustrates the advantage of using ViTA to resolve ambiguities. mBART is translating the word court to judicial court, while ViTA translates it to tennis court.}
    \label{fig:improvement_eg}
\end{figure}

It can be seen that our model is able to generalize well on the challenge set as well and performs better than other systems by a large margin.
To further analyze the results, we find a few cases in the challenge set wherein ViTA is able to resolve ambiguities, and an example is illustrated in Figure \ref{fig:improvement_eg}.
Yet, the performance of the models is very similar across the textual-only and multimodal domains, and there are no significant improvements observed in the multimodal system.

\subsection{Degradation}

Although there is no significant improvement in the multimodal systems over the textual-only models, \citet{caglayan-etal-2019-probing} explore the robustness of multimodal systems by systematically degrading the source text for translations. We employ a similar approach and degrade the source text to compare our systems.

\subsubsection{Entity masking}

\begin{figure}
    \centering
    \includegraphics[width=\columnwidth,keepaspectratio]{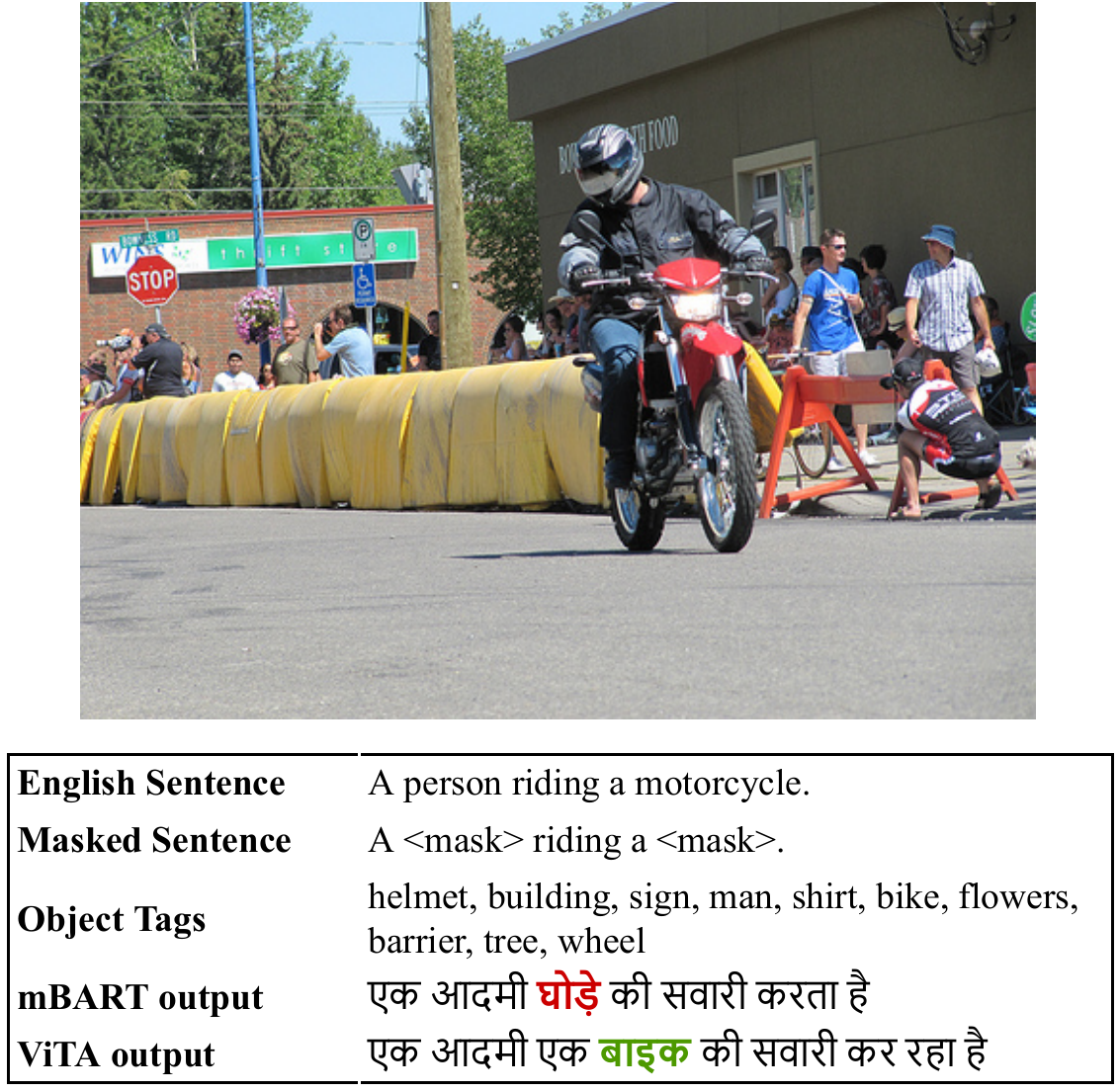}
    \caption{The effect of object tags on an entity masked input from the test set. ViTA is able to use the context built from the object tags to predict a motorcycle, while mBART is predicting a horse instead.}
    \label{fig:entity_example}
\end{figure}

\begin{table}
\centering
\resizebox{\columnwidth}{!}{%
\begin{tabular}{lrrrr}
\toprule
& \textbf{Train} & \textbf{Valid} & \textbf{Test} & \textbf{Challenge} \\ \midrule
\#entities in text & 29,583 & 1,028 & 1,631 & 1,592 \\
\#objects tags in images & 253,051 & 8,679 & 13,855 & 12,507 \\
\#entities in object tags & 13,959 & 498 & 758 & 442 \\
\%entities in object tags & 47.18\% & 48.44\% & 46.47\% & 27.76\% \\
\bottomrule
\end{tabular}%
}
\caption {We show the overlap between the entities in the text and the object tags detected using Faster R-CNN model. The entities were identified using the \texttt{en\_core\_web\_sm} model from the spaCy library\textsuperscript{\ref{spacyref}}.}
\label{tab:overlap}
\end{table}

\begin{figure*}
\begin{minipage}{.5\linewidth}
\centering
\subfloat[\centering Entity Masking]{\label{main:a}\includegraphics[scale=.5,keepaspectratio]{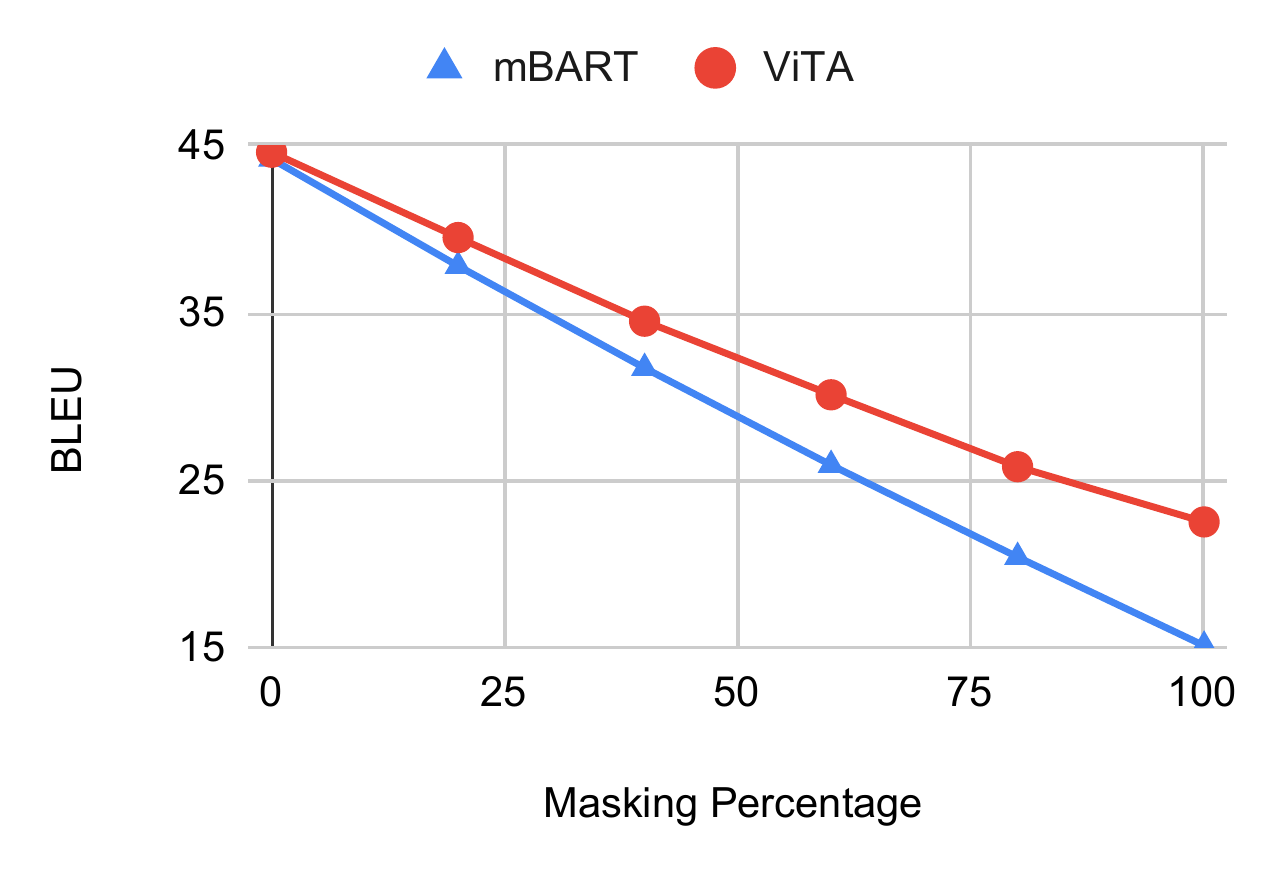}}
\end{minipage}%
\begin{minipage}{.5\linewidth}
\centering
\subfloat[\centering Color Deprivation]{\label{main:b}\includegraphics[scale=.5,keepaspectratio]{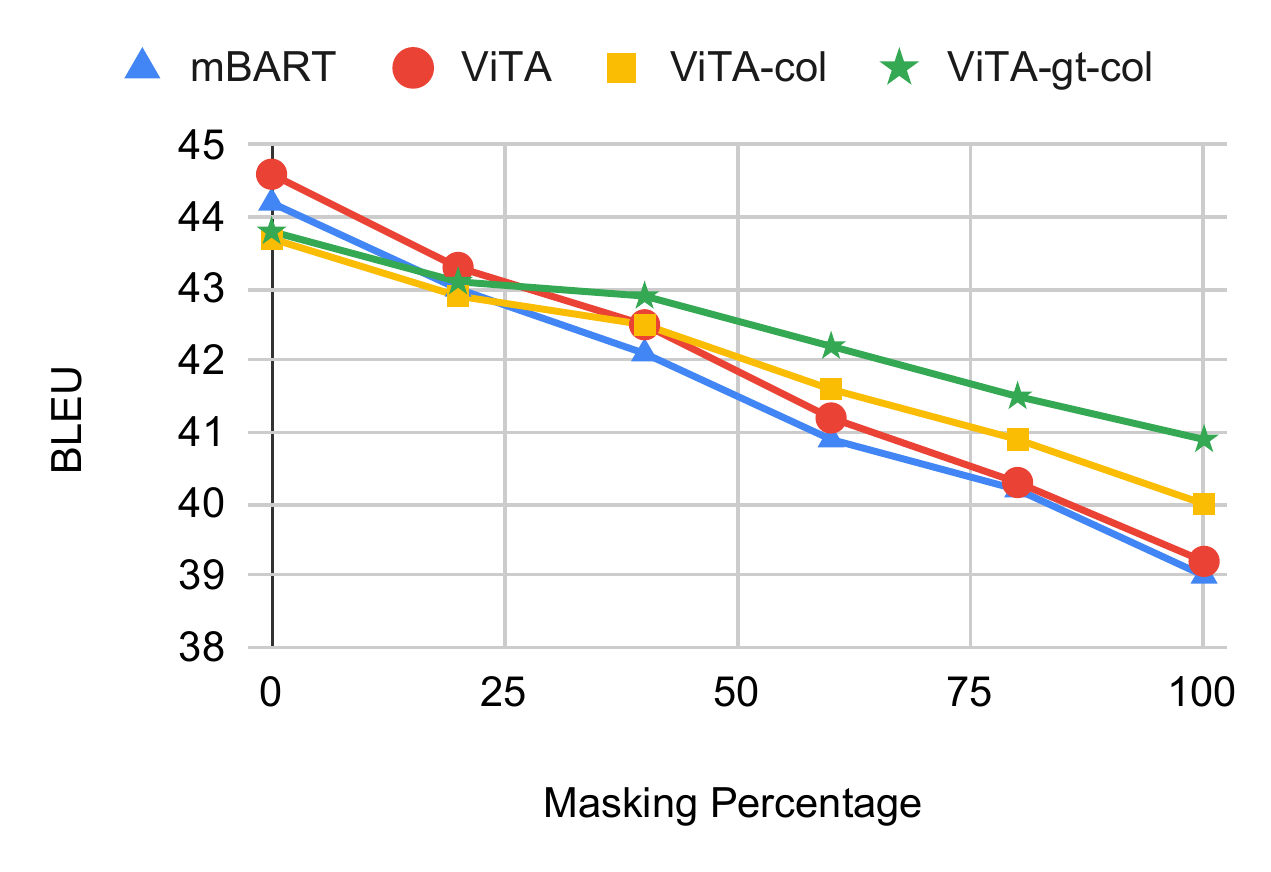}}
\end{minipage}\par\medskip
\begin{minipage}{.5\linewidth}
\centering
\subfloat[\centering Adjective Masking]{\label{main:c}\includegraphics[scale=.5,keepaspectratio]{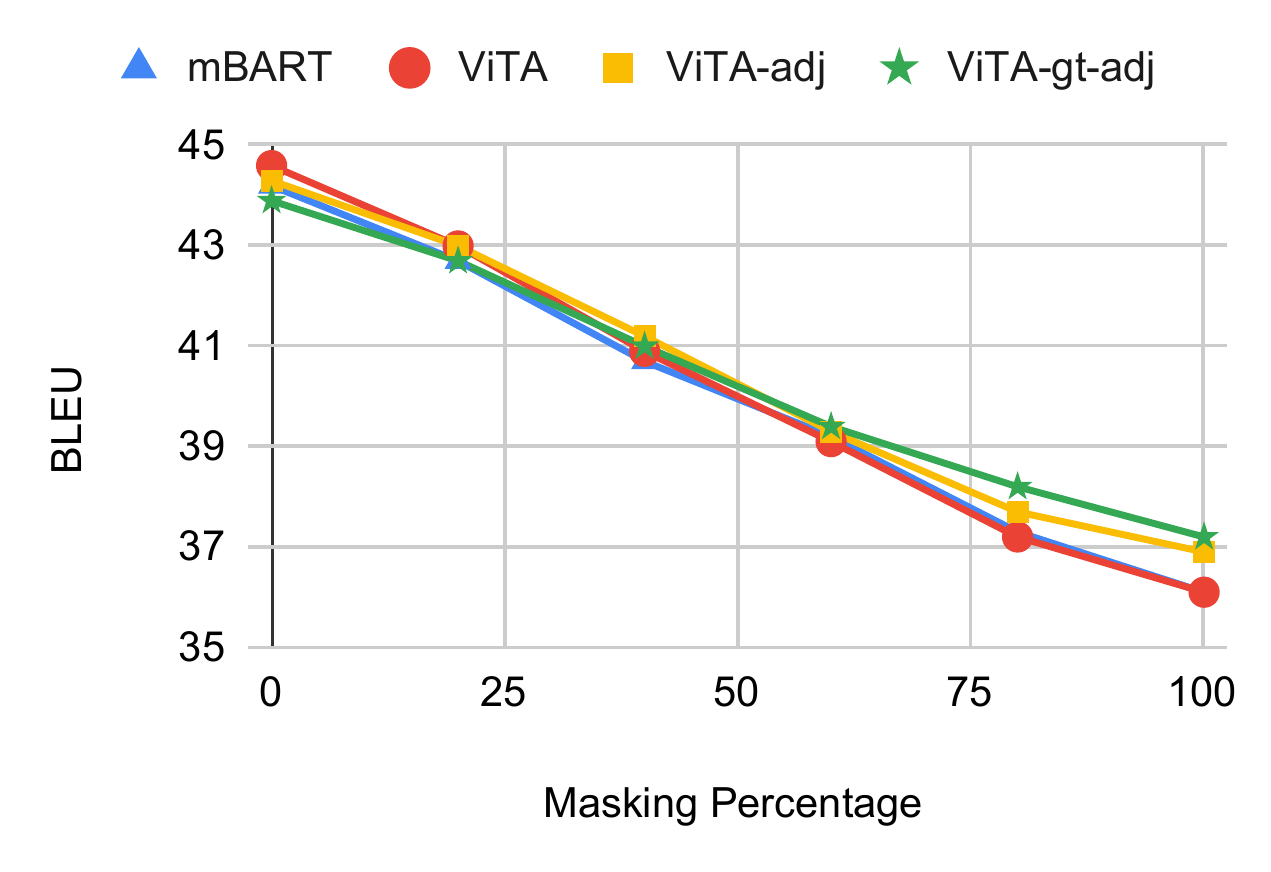}}
\end{minipage}%
\begin{minipage}{.5\linewidth}
\centering
\subfloat[\centering Random Masking]{\label{main:d}\includegraphics[scale=.5,keepaspectratio]{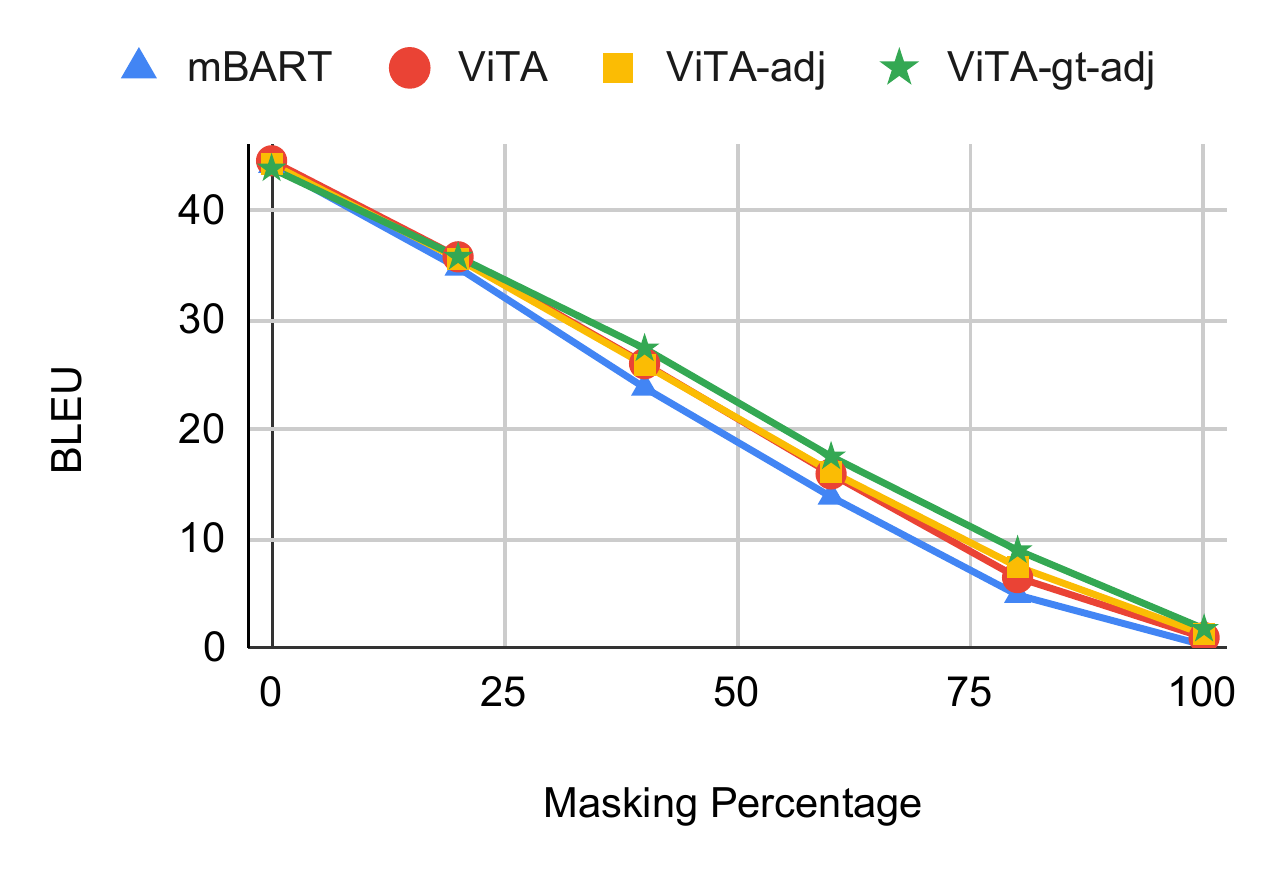}}
\end{minipage}

    \caption{BLEU score comparison of the proposed models by increasing the masking percentage in the source text.}
    \label{fig:degradation}%

\end{figure*}



\begin{table}[t]
\centering
\resizebox{\columnwidth}{!}{
\begin{tabular}{lrrr}
\toprule
 & \textbf{No masking} & \textbf{Entity Masking} & \textbf{Degradation \%}  \\ \midrule

mBART & 44.2 & 15.1 & 65.8 \\
ViTA & \textbf{44.6} & 22.5 & 49.6 \\ 
ViTA-gt & 43.6 & \textbf{25.4} & \textbf{41.7} \\

\bottomrule
\end{tabular}
}
\caption{The effect of entity masking on the BLEU score of the proposed models on the test set.}
\label{tab:entity}
\end{table}

The goal of entity masking is to mask out the visually depictable entities in the source text so that the multimodal systems can make use of the visual cues in the image. To identify such entities, we use the \texttt{en\_core\_web\_sm} model in spaCy\footnote{\label{spacyref}\url{https://spacy.io/}} to predict the nouns in the sentence. The statistics of the tagged entities can be seen in Table \ref{tab:overlap}.

We progressively increase the percentage of masked entities to better compare the degradation of our systems and it can be seen in Figure \ref{main:a}.
The final degraded values are reported in Table \ref{tab:entity}. Since the masked entities can also be predicted by using only the textual context of the sentence, we similarly add a training step of masking \approximately15\% tokens while training mBART for a valid comparison. An example of the performance of our systems on an entity masked input is illustrated in Figure \ref{fig:entity_example}.

As an upper bound to the scope of our system, we propose \texttt{ViTA-gt}, which uses the ground-truth object labels from the Visual Genome dataset. Since the number of annotated objects is large, we filter them by removing the objects far from the image region.

\begin{table}[t]
\centering
\resizebox{\columnwidth}{!}{
\begin{tabular}{lrrr}
\toprule
 & \textbf{No masking} & \textbf{Color Deprivation} & \textbf{Degradation \%}  \\ \midrule
mBART & 44.2 & 39.0 & 11.8 \\
ViTA & \textbf{44.6} & 39.2 & 12.1 \\
ViTA-col & 43.7 & 40.0 & 8.5 \\
ViTA-gt-col & 43.8 & \textbf{40.9} & \textbf{6.6} \\
\bottomrule
\end{tabular}
}
\caption{The effect of color deprivation on the BLEU score of the proposed models on the test set.}
\label{tab:color}
\end{table}

\subsubsection{Color deprivation}

The goal of color deprivation is to similarly mask tokens that are difficult to predict without the visual context of the image.
To identify the colors in the source text, we maintain a list of colors and check whether the words in the sentence are present in the list.
Similar to entity masking, we progressively increase the percentage of masked colors in the dataset to compare our systems. The comparison of our systems can be seen in Figure \ref{main:b}. The final values of color deprivation are reported in Table \ref{tab:color}.

As an upper bound to the scope of our system, we believe that colors can further be added to the object tags to help build a more robust system. As an added experiment, we propose \texttt{ViTA-col} by using the ground-truth annotations from the Visual Genome dataset and adding colors to our predicted object tags, which are present in the ground-truth objects as well.
As a part of future work, we would like to extend our system to predict the colors from the image itself.
We further experiment with \texttt{ViTA-gt-col}, which uses ground-truth objects with added colors in the input.

\subsubsection{Adjective Masking}

\begin{table}[t]
\centering
\resizebox{\columnwidth}{!}{%
\begin{tabular}{lrrr}
\toprule
            & \textbf{No masking} & \textbf{Adjective Masking} & \textbf{Degradation \%}  \\ \midrule

mBART & 44.2 & 36.1 & 18.3 \\
ViTA & \textbf{44.6} & 36.1 & 19.1 \\
ViTA-adj & 44.3 & 36.9 & 16.7 \\
ViTA-gt-adj & 43.9 & \textbf{37.2} & \textbf{15.3} \\

\bottomrule
\end{tabular}
}
\caption{The effect of adjective masking on the BLEU score of the proposed models on the test set.}
\label{tab:adjective}
\end{table}

Similar to color deprivation, we propose adjective masking as several of the adjectives are visually depictable, and the degradation comparison should not be limited to just entities and colors. We predict the adjectives in the sentence by using the POS tagging model \texttt{en\_core\_web\_sm} from spaCy library.

The performance of our models is compared in Figure \ref{main:c}. The final values are reported in Table \ref{tab:adjective}.

As an upper bound to the scope of our system, we propose to add all the adjectives to their corresponding object tags in the input. We propose \texttt{ViTA-adj} by adding the ground truth adjectives annotated in the Visual Genome dataset to the object tags which are also predicted by our object detector. We also propose \texttt{ViTA-gt-adj}, which uses the ground-truth objects with their corresponding adjectives. The objects which are from the image region are removed to mitigate the noise added by the large number of objects in the annotations.

\subsubsection{Random Masking}

For a general robustness comparison of our models, we remove the limitation of manually masking the source sentences and progressively mask the text by random sampling.

The performance of our models is compared in Figure \ref{main:d}.

\section{Conclusion}

We propose a multimodal translation system and utilize the textual-only pre-training of a neural machine translation system, mBART, by extracting object tags from the image. Further, we explore the robustness of our proposed multimodal system by systematically degrading the source texts and observe improvements from the textual-only counterpart. We also explore the shortcomings of the currently available object detectors and use ground-truth annotations in our experiments to show the scope of our methodology. The addition of colors and adjectives further adds to the robustness of the system and can be explored further in the future.



\bibliographystyle{acl_natbib}
\bibliography{anthology,acl2021}


\end{document}